# Sequence Tutor: Conservative Fine-Tuning of Sequence Generation Models with KL-control

Natasha Jaques [1 2]  Shixiang Gu [1 3 4]  Dzmitry Bahdanau [1 5]  José Miguel Hernández-Lobato [3]
Richard E. Turner [3]  Douglas Eck [1]

## Abstract

This paper proposes a general method for improving the structure and quality of sequences generated by a recurrent neural network (RNN), while maintaining information originally learned from data, as well as sample diversity. An RNN is first pre-trained on data using maximum likelihood estimation (MLE), and the probability distribution over the next token in the sequence learned by this model is treated as a prior policy. Another RNN is then trained using reinforcement learning (RL) to generate higher-quality outputs that account for domain-specific incentives while retaining proximity to the prior policy of the MLE RNN. To formalize this objective, we derive novel off-policy RL methods for RNNs from KL-control. The effectiveness of the approach is demonstrated on two applications; 1) generating novel musical melodies, and 2) computational molecular generation. For both problems, we show that the proposed method improves the desired properties and structure of the generated sequences, while maintaining information learned from data.

## 1. Introduction

The approach of training sequence generation models using likelihood maximization suffers from known failure modes, and it is notoriously difficult to ensure multi-step generated sequences have coherent global structure. For example, long short-term memory (LSTM) (Hochreiter & Schmidhuber, 1997) networks trained to predict the next character in sequences of text may produce text that has correct spelling, punctuation, and even a semblance of grammar, but the generated text shifts so rapidly from topic to topic, that it is almost completely nonsensical (see (Graves, 2013) for an example). Similar networks trained to predict the next note in a melody suffer from the same problem; the generated music has no consistent theme or structure, and appears wandering and random. In addition, these models are prone to excessively repeating the same output token, a problem that has also been noted in the context of recurrent dialog generation models (Li et al., 2016).

To ameliorate these problems we propose *Sequence Tutor*, a novel approach which uses RL to impose structure on a sequence generation RNN via task-specific rewards, while simultaneously ensuring that information learned from data is retained. This is accomplished by maintaining a fixed copy of a sequence generation RNN pre-trained on data, which is termed the *Reward RNN*. Rather than simply using the Reward RNN to supply part of the rewards to our model, we derive novel off-policy RL methods for sequence generation from KL-control that allow us to directly penalize Kullback Leibler (KL) divergence from the policy defined by the Reward RNN. As a byproduct of minimizing KL our objective includes an entropy regularization term that encourages high entropy in the distribution of the RL model. This is ideal for sequence generation tasks such as text, music, or molecule generation, in which maintaining diversity in the samples generated by the model is critical.

Sequence Tutor effectively combines both data and task-related goals, without relying on either as a perfect metric of task success. This is an important novel direction of research. Much previous work on combining RL and MLE has used MLE training simply as a way to bootstrap the training of an RL model (Ranzato et al., 2015; Bahdanau et al., 2016; Li et al., 2016), since training with RL from scratch is difficult. However, this approach does not encourage diversity of the generated samples, and can be problematic when task-specific rewards are incomplete or imperfect. Designing an appropriate reward definition is highly non-trivial, and often the hand-crafted rewards cannot be fully trusted (Vedantam et al., 2015; Liu et al., 2016). And yet, relying on data alone can be insufficient when the data itself contains biases, as has been shown for text data

[1]Google Brain, Mountain View, USA [2]Massachusetts Institute of Technology, Cambridge, USA [3]University of Cambridge, Cambridge, UK [4]Max Planck Institute for Intelligent Systems, Stuttgart, Germany [5]Université de Montréal, Montréal, Canada. Correspondence to: Natasha Jaques <jaquesn@mit.edu>.

(Caliskan-Islam et al., 2016), or when domain-specific constraints cannot be encoded directly into MLE training. By learning a policy that trades off staying close to the data distribution while improving performance on specific metrics, Sequence Tutor reduces both of these problems.

This paper contributes to the sequence training and RL literature by a) proposing a novel method for combining MLE and RL training; b) showing the connection between KL control and sequence generation; c) deriving the explicit relationships among a generalized variant of $\Psi$-learning (Rawlik et al., 2012), G-learning (Fox et al., 2015), and Q-learning with log prior augmentation, and being the first to empirically compare these methods and use them with deep neural networks.

We explore the usefulness of our approach for two sequence generation applications. The first, music generation, is a difficult problem in which the aesthetic beauty of generated sequences cannot be fully captured in a known reward function, but in which models trained purely on data cannot produce well-structured sequences. Through an empirical study, we show that by imposing rules of music theory on a melody generation model, Sequence Tutor is able to produce melodies which are varied, yet more harmonious, interesting, and rated as significantly more subjectively pleasing than those of the MLE model. Further, Sequence Tutor is able to significantly reduce unwanted behaviors and failure modes of the original RNN. The effectiveness of Sequence Tutor is also demonstrated for computational molecular generation, a task in which the goal is to generate novel drug-like molecules with desirable properties by outputting a string representation of the molecule encoding. However, generating valid molecules can prove difficult, as it is hard for probabilistic models to learn all the constraints that define physically realizable molecules directly from data (Gómez-Bombarelli et al., 2016). We show that Sequence Tutor is able to yield a higher percentage of valid molecules than the baseline MLE RNN, and the generated molecules score higher on metrics of drug-likeness and ease of synthesis.

## 2. Related Work

Recent work has attempted to use both MLE and RL in the context of structured prediction. While the attempts were successful, the problems of maintaining information about the data distribution and diversity in the generated samples were not addressed. MIXER (Mixed Incremental Cross-Entropy Reinforce) (Ranzato et al., 2015) uses BLEU score as a reward signal to gradually introduce a RL loss to a text translation model. Bahdanau et al. (2016) applies an actor-critic method and uses BLEU score directly to train a critic network to output the value of each word, where the actor is again initialized with the policy of an RNN trained with next-step prediction. Li et al. (2016) use RL to improve a pre-trained dialog model with heuristic rewards. These approaches assume that the complete task reward specification is available. They pre-train a good policy with supervised learning so that RL can be used to learn the true task objective, since it can be difficult to reach convergence when training with pure RL. However, the original MLE policy of these models is overwritten by the RL training process. In contrast, Sequence Tutor uses rewards to correct certain properties of the generated data, while learning most information from data and maintaining this information; an important ability when the true reward function is not available or imperfect.

Reward augmented maximum likelihood (RAML) (Norouzi et al., 2016) is an approach designed to improve MLE training of a translation model by augmenting the ground truth targets with additional outputs that are within a small edit distance, and performing MLE training against those as well. The authors show that their approach is equivalent to minimizing KL-divergence between an RL *exponentiated payoff* distribution based on edit distance, and the MLE distribution. In contrast, our goal is generation rather than prediction, and we train an RL rather than MLE model. The RAML approach, while an important contribution, is only viable if it is possible to generate additional MLE training samples that are similar in terms of the reward function to the ground truth (i.e. samples within a small edit distance). However in some domains, including the two explored in this paper, generating similar samples with high reward is not only not possible, but in fact constitutes the entire problem under investigation.

Finally, our approach is related to KL control (Todorov, 2007; Kappen et al., 2012; Rawlik et al., 2012), a branch of stochastic optimal control (SOC) (Stengel, 1986). There is also a connection between this work and Maximum Entropy Inverse RL (Ziebart et al., 2008), which can be seen as KL control with a flat, improper prior. From KL control, we take inspiration from two off-policy, model-free methods, $\Psi$-learning (Rawlik et al., 2012) and $G$-learning (Fox et al., 2015). Both approaches are derived from a KL-regularized RL objective, where an agent maximizes the reward while incurring additional penalty for divergence from some prior policy. While our methods rely on similar derivations presented in these papers, our methods have different motivations and forms from the original papers. The original $\Psi$-learning (Rawlik et al., 2012) restricts the prior policy to be the policy at the previous iteration and solves the original RL objective with conservative, KL-regularized policy updates, similar to conservative policy gradient methods (?Peters et al., 2010; Schulman et al., 2015). The original $G$-learning (Fox et al., 2015) penalizes divergence from a simple uniform prior policy in order to cope with over-estimation of target $Q$ values. These tech-niques have not been applied to deep learning techniques or with RNNs, or as a way to improve a pre-trained MLE model. Our work is the first to explore these methods in such a context, and includes a $Q$-learning model with additional cross-entropy reward as a comparable alternative. To the best of our knowledge, our work is the first to provide comparisons among these three approaches.

There has also been prior work in the domain of generative modeling of music. Using RNNs for this purpose has been explored in a variety of contexts, including generating Celtic folk music (Sturm et al., 2016), or improvising the blues (Eck & Schmidhuber, 2002). Often, this involves training the RNN to predict the next note in a monophonic melody; however, as mentioned above, the melodies generated by this model tend to wander and lack musical structure. Some authors have experimented with encoding musical structure into a hierarchical RNN with layers dedicated to generated the melody, drums, and chords (Chu et al., 2016). Other approaches have examined RNNs with richer expressivity, latent-variables for notes, or raw audio synthesis (Boulanger-Lewandowski et al., 2012; Gu et al., 2015; Chung et al., 2015). Recently, *Wavenet* produced impressive performance in generating music from raw audio using convolutional neural networks with receptive fields at various time scales (van den Oord et al., 2016). However, the authors themselves note that "even with a receptive field of several seconds, the models did not enforce long-range consistency which resulted in second-to-second variations in genre, instrumentation, and sound quality" (p. 8).

Finally, prior work has successfully performed computational molecular generation with deep neural networks. Segler et al. (2017) demonstrated that an LSTM trained on sets of biologically active molecules can be used to generate novel molecules with similar properties. Gómez-Bombarelli et al. (2016) trained a variational autoencoder to learn a compact embedding of molecules encoded using the SMILES notation. By interpolating in the embedding space and optimizing for desirable metrics of drug quality, the authors were able to decode molecules with high scores on these metrics. However, producing embeddings that led to valid molecules was difficult; in some cases, as little as 1% of generated sequences proved to be a valid molecule encoding.

## 3. Background

In RL, an agent interacts with an environment. Given the state of the environment at time $t$, $s_t$, the agent takes an action $a_t$ according to its policy $\pi(a_t|s_t)$, receives a reward $r(s_t, a_t)$, and the environment transitions to state, $s_{t+1}$. The agent's goal is to maximize reward over a sequence of actions, with a discount factor of $\gamma$ applied to future rewards. The optimal deterministic policy $\pi^*$ is known to satisfy the following Bellman optimality equation,

$$Q(s_t, a_t; \pi^*) = r(s_t, a_t) \qquad (1)$$
$$+ \gamma \mathbb{E}_{p(s_{t+1}|s_t, a_t)}[\max_{a_{t+1}} Q(s_{t+1}, a_{t+1}; \pi^*)]$$

where $Q^\pi(s_t, a_t) = \mathbb{E}_\pi[\sum_{t'=t}^{\infty} \gamma^{t'-t} r(s_{t'}, a_{t'})]$ is the $Q$ function of a policy $\pi$. In Deep $Q$-learning (Mnih et al., 2013), a neural network called the deep Q-network (DQN) is trained to approximate $Q(s, a; \theta)$, using the following objective,

$$L(\theta) = \mathbb{E}_\beta[(r(s, a) + \gamma \max_{a'} Q(s', a'; \theta^-) - Q(s, a; \theta))^2] \qquad (2)$$

where $\beta$ is the exploration policy, and $\theta^-$ is the parameters of the target Q-network (Mnih et al., 2013) that is held fixed during the gradient computation. The target Q-network is updated more slowly than the $Q$-network; for example the moving average of $\theta$ can be used as $\theta^-$, as proposed by Lillicrap et al. (2015). Exploration can be performed with either the $\epsilon$-greedy method or Boltzmann sampling. Additional techniques such as a *replay memory* (Mnih et al., 2013) are used to stabilize and improve learning.

## 4. Sequence Tutor

Given a trained sequence generation RNN, we would like to impose domain-specific rewards based on the structure and quality of generated sequences, while still maintaining information about typical sequences learned from data. Therefore, we treat the trained model as a black-box prior policy, and focus on developing a method that can tune some properties of the model without interfering with the original probability distribution learned from data. The separation between the trained sequence model and the tuning method is important, as it prevents RL training from overwriting the original policy. To accomplish this task, we propose Sequence Tutor. An LSTM trained on data supplies the initial weights for three networks in the model: a recurrent $Q$-network and target $Q$-network, and a Reward RNN. The Reward RNN is held fixed during training, and treated as a prior policy which can supply the probability of a given token in a sequence as originally learned from data.

To apply RL to sequence generation, generating the next token in the sequence is treated as an action $a$. The state of the environment consists of all of the tokens generated so far, i.e. $s_t = \{a_1, a_2, ...a_{t-1}\}$. Given action $a_t$, we would like the reward $r_t$ to combine information about the prior policy $p(a_t|s_t)$ as output by the Reward RNN, as well as some domain- or task-specific rewards $r_T$. Figure 1 illustrates these ideas.

Sequence Tutor: Conservative Fine-Tuning of Sequence Generation Models with KL-control

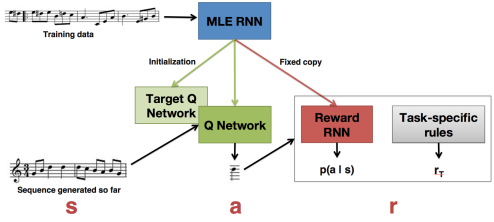

Figure 1: An RNN pre-trained on data using MLE supplies the initial weights for the $Q$-network and target $Q$-network, and a fixed copy is used as the Reward RNN.

### 4.1. Q-learning with log prior augmentation

The simplest and most naïve way to incorporate information about the prior policy is to directly augment the task-specific rewards with the output of the Reward RNN. In this case, the total reward given at time $t$ becomes:

$$r(s,a) = \log p(a|s) + r_T(a,s)/c \qquad (3)$$

where $c$ is a constant controlling the emphasis placed on the task-specific rewards. Given the DQN objective in Eq. 2 and modified reward function in Eq. 3, the objective and learned policy are:

$$L(\theta) = \mathbb{E}_\beta[(\log p(a|s) + r_{MT}(a,s)/c \qquad (4)$$
$$+ \gamma \max_{a'} Q(s',a';\theta^-) - Q(s,a;\theta))^2]$$

$$\pi_\theta(a|s) = \delta(a = \arg\max_a Q(s,a;\theta)). \qquad (5)$$

This modified objective forces the model to learn that the most valuable actions are those that conform to the music theory rules, but still have high probability in the original data. However, the DQN learns a deterministic policy (as shown in Eq. 5), which is not ideal for sequence generation. Therefore, after the model is trained, we generate sequences by sampling from the softmax function applied to the predicted $Q$-values.

### 4.2. KL Control for Sequence Generation

If we cast sequence generation as a sequential decision-making problem and the desired sequence properties in terms of target rewards, the problem can be expressed as a KL control problem for a non-Markovian system. KL control (Todorov, 2007; Kappen et al., 2012; Rawlik et al., 2012) is a branch of stochastic optimal control (SOC) (Stengel, 1986), which studies an RL, or control, problem in which the agent tries maximizing its task reward while minimizing deviation from a prior policy. For our purposes, we treat a trained MLE sequence model as the prior policy, and thus the objective is to train a new policy, or sequence model, to maximize some rewards while keeping close to the original MLE model. We show that such KL control formulation allows us to derive additional variants of Q-learning with minimal modifications, which give rise to different properties. Let $\tau = \{a_1, a_2, ..., a_{t-1}\}$ represent the sequence, $r(\tau)$ the reward of the sequence, $p(\tau)$ be the prior distribution over $\tau$ given by the trained sequence model, and $q(\tau)$ be the policy of the Sequence Tutor model. The objective is then to maximize the following expression with respect to $q(\tau)$, where $D_{KL}$ represents the KL divergence of distributions:

$$L(q) = \mathbb{E}_{q(\tau)}[r(\tau)]/c - D_{KL}[q(\tau)||p(\tau)]. \qquad (6)$$

We express $q(\tau)$ in terms of a parametrized recurrent policy $\pi_\theta(a_t|s_t)$, i.e. $q(\tau) = \prod_{t=1}^T \pi_\theta(a_t|s_t)$ where $s_t = \{a_1, a_2, ..., a_{t-1}\}$, indicates that the system is non-Markovian. The prior policy is expressed similarly $p(\tau) = \prod_{t=1}^T p(a_t|s_t)$. The reinforcement learning objective is the following, where $\mathbb{E}_\pi[\cdot]$ below indicates expectation with respect to sequences sampled from $\pi$,

$$L(\theta) = \mathbb{E}_\pi[\sum_t r(s_t, a_t)/c + \log p(a_t|s_t) - \log \pi_\theta(a_t|s_t)]$$

The difference between this equation and Eq. 4 is that an entropy regularizer is now included, and thus the optimal policy is no longer deterministic. Below, we derive general temporal-difference based methods for the KL-control problem for sequence generation.

### 4.3. Recurrent Generalized $\Psi$-learning

Let $V^\pi(s_t)$ define the recurrent value function of the policy $\pi_\theta$, given by,

$$V^\pi(s_t) = \mathbb{E}_\pi[\sum_{t'=t}^\infty r(s_{t'}, a_{t'})/c + \log p(a_{t'}|s_{t'}) \qquad (7)$$
$$- \log \pi(a_{t'}|s_{t'})]$$

We define the generalized $\Psi$ function, analogous to $Q$ function for KL control, as below. We call this generalized $\Psi$ function, as it was introduced in deriving $\Psi$-learning (Rawlik et al., 2012), and the following derivation is a generalization to the $\Psi$-learning algorithm.

$$\Psi^\pi(s_t, a_t) = r(s_t, a_t)/c + \log p(a_t|s_t) + V^\pi(s_{t+1}) \qquad (8)$$

Note that the state $s_{t+1}$ is given deterministically by $s_t = \{a_1, a_2, ..., a_{t-1}\}$ and $a_t$ for sequence modeling, and thus the expressions do not contain the usual stochastic dynamics $p(s_{t+1}|s_t, a_t)$. The value function $V^\pi(s_t)$ can be recursively expressed in terms of $\Psi^\pi$,

$$V^\pi(s_t) = \mathbb{E}_\pi[\Psi^\pi(s_t, a_t)] + \mathbb{H}[\pi(.|s_t)] \qquad (9)$$
$$= \mathbb{E}_\pi[\Psi^\pi(s_t, a_t) - \log \pi(a_t|s_t)] \qquad (10)$$

Fixing $\Psi(s_t, a_t) = \Psi^\pi(s_t, a_t)$ and constraining $\pi$ to be a probability distribution, the optimal greedy policy update

$\pi^*$ can be derived, along with the corresponding optimal value function,

$$\pi^*(a_t|s_t) \propto e^{\Psi(s_t,a_t)} \tag{11}$$

$$V^*(s_t) = \log \sum_{a_t} e^{\Psi(s_t,a_t)} \tag{12}$$

Given Eq. 8 and 12, the following Bellman optimality equation for generalized $\Psi$ function is derived.

$$\Psi^*(s_t, a_t) = r(s_t, a_t)/c + \log p(a_t|s_t) \\ + \log \sum_{a_{t+1}} \exp(\Psi^*(s_{t+1}, a_{t+1})) \tag{13}$$

The $\Psi$-learning loss directly follows:

$$L_\Psi(\theta) = \mathbb{E}_\beta[(\Psi_\theta(s_t, a_t) - y_t)^2] \text{ where} \tag{14}$$
$$y_t = \log p(a_t|s_t) + r(s_t, a_t)/c + \gamma \log \sum_{a'} e^{\Psi^-(s_{t+1}, a')}$$

$\beta$ corresponds to sampling sequence trajectories from an arbitrary distribution; in practice, the experience replay (Mnih et al., 2013). $\Psi^-$ indicates that it uses the target network. $\Psi_\theta$, i.e. $\pi_\theta$, is parametrized with recurrent neural networks, and for discrete actions, $\pi_\theta$ is effectively a softmax layer on top of $\Psi_\theta$.

### 4.4. Recurrent $G$-learning

We can derive another algorithm by parametrizing $\Psi_\theta$ indirectly by $\Psi_\theta(s_t, a_t) = \log p(a_t|s_t) + G_\theta(s_t, a_t)$. Substituting into above equations, we get a different temporal-difference method:

$$L_G(\theta) = \mathbb{E}_\beta[(G_\theta(s_t, a_t) - y_t)^2] \text{ where} \tag{15}$$
$$y_t = r(s_t, a_t)/c + \gamma \log \sum_{a'} p(a'|s_{t+1}) e^{G^-(s_{t+1}, a')} \text{ and}$$
$$\pi_\theta(a_t|s_t) \propto p(a_t|s_t) \exp(G_\theta(s_t, a_t))$$

This formulation corresponds to $G$-learning (Fox et al., 2015), which can thus be seen as a special case of generalized $\Psi$-learning. Unlike $\Psi$ learning, which directly builds knowledge about the prior policy into the $\Psi$ function, the $G$-function does not give the policy directly but instead needs to be dynamically mixed with the prior policy probabilities. While this computation is straight-forward for discrete action domains as here, extensions to continuous action domains require additional considerations such as normalizability of $\Psi$-function parametrization (Gu et al., 2016).

The KL control-based derivation also has another benefit in that the stochastic policies can be directly used as an exploration strategy, instead of heuristics such as $\epsilon$-greedy or additive noise (Mnih et al., 2013; Lillicrap et al., 2015).

### 4.5. Sequence Tutor implementation

Following from the above derivations, we compare three methods for implementing Sequence Tutor: $Q$-learning with log prior augmentation (based on Eq. 4), generalized $\Psi$-learning (based on Eq. 14), and $G$-learning (based on Eq. 15). A pre-trained sequence generation LSTM is used as the Reward RNN, to supply the cross entropy reward in $Q$-learning and the prior policy in $G$- and generalized $\Psi$-learning. These approaches are compared to both the original performance of the MLE RNN, and a model trained using only RL and no prior policy. Model evaluation is performed every 100,000 training epochs, by generating 100 sequences and assessing the average $r_T$ and $\log p(a|s)$. The code for Sequence Tutor, including a checkpointed version of the trained melody RNN is available at https://github.com/tensorflow/magenta/tree/master/magenta/models/rl_tuner.

## 5. Experiment I: Melody Generation

Music compositions adhere to relatively well-defined structural rules, making music an interesting sequence generation challenge. For example, music theory tells that groups of notes belong to keys, chords follow progressions, and songs have consistent structures made up of musical phrases. Our research question is therefore whether such constraints can be learned by an RNN, while still allowing it to maintain note probabilities learned from data.

To test this hypothesis, we developed several rules that we believe describe pleasant-sounding melodies, taking inspiration from a text on melodic composition (Gauldin, 1995). We do not claim these characteristics are exhaustive or strictly necessary for good composition; rather, they are an incomplete measure of task success that can simply guide the model towards traditional composition structure. It is therefore crucial that the Sequence Tutor approach allows the model to retain knowledge learned from real songs in the training data. The rules comprising the music-specific reward function $r_T(a, s)$ encourage melodies to: stay in key, start with the tonic note, resolve melodic leaps, have a unique maximum and minimum note, prefer harmonious intervals, play motifs and repeat them, have a low autocorrelation at a lag of 1, 2, and 3 beats, and avoid excessively repeating notes. Interestingly, while excessively repeating tokens is a common problem in RNN sequence generation models, avoiding this behavior is also Gauldin's first rule of melodic composition (p. 42).

To train the model, we begin by extracting monophonic melodies from a corpus of 30,000 MIDI songs and encoding them as one-hot sequences of notes[1]. These melodies are then used to train an LSTM with one layer of 100 cells. Optimization was performed with Adam (Kingma & Ba, 2014), a batch size of 128, initial learning rate of .5, and a stepwise learning rate decay of 0.85 every 1000 steps. Gradients were clipped to ensure the L2 norm was less than 5, and weight regularization was applied with $\beta = 2.5 \times 10^{-5}$. Finally, the losses for the first 8 notes of each sequence were not used to train the model, since it cannot reasonably be expected to accurately predict them with no context. The trained RNN eventually obtained a validation accuracy of 92% and a log perplexity score of .2536. This model was used as described above to initialize the three sub-networks in the Sequence Tutor model.

---
[1] More information about both the note encoding and the reward metrics is available in the supplementary material.

The Sequence Tutor model was trained using a similar configuration to the one above, except with a batch size of 32, and a reward discount factor of $\gamma$=.5. The Target-$Q$-network's weights $\theta^-$ were gradually updated towards those of the $Q$-network ($\theta$) according to the formula $(1-\eta)\theta^- + \eta\theta$, where $\eta = .01$ is the Target-$Q$-network update rate. A strength of our model is that the influence of data and task-specific rewards can be explicitly controlled by adjusting the temperature parameter $c$. We replicated our results for a number of settings for $c$; we present results for $c$=.5 below because we believe them to be most musically pleasing, however additional results are available at https://goo.gl/cTZy8r. Similarly, we replicated the results using both $\epsilon$-greedy and Boltzmann exploration, and present the results using $\epsilon$-greedy exploration below.

### 5.1. Results

Table 1 provides quantitative results in the form of performance on the music theory rules to which we trained the model to adhere; for example, we can assess the fraction of notes played by the model which belonged to the correct key, or the fraction of melodic leaps that were resolved. The statistics were computed by randomly generating 100,000 melodies from each model.

The results above demonstrate that the application of RL is able to correct almost all of the targeted "bad behaviors" of the MLE RNN, while improving performance on the desired metrics. For example, the original LSTM model was extremely prone to repeating the same note; after applying RL, we see that the number of notes belonging to some excessively repeated segment has dropped from 63% to nearly 0% in all of the Sequence Tutor models. While the metrics for the G model did not improve as consistently, the $Q$ and $\Psi$ models successfully learned to adhere to most of the imposed rules. The degree of improvement on these metrics is related to the magnitude of the reward given for the behavior. For example, a strong penalty of -100 was applied each time a note was excessively repeated, while a reward of only 3 was applied at the end of a melody for unique extrema notes (which most likely explains the lack of improvement on this metric). The reward values could be adjusted to improve the metrics further, however we found that these values produced pleasant melodies.

| Metric | MLE | Q | $\Psi$ | G |
|---|---|---|---|---|
| Repeated notes | 63.3% | **0.0%** | **0.02%** | **0.03%** |
| Mean autocorr. lag 1 | -.16 | **-.11** | **-.10** | .55 |
| Mean autocorr. lag 2 | .14 | **.03** | **-.01** | .31 |
| Mean autocorr. lag 3 | -.13 | **.03** | **.01** | 17 |
| Notes not in key | 0.1% | 1.00% | 0.60% | 28.7% |
| Starts with tonic | 0.9% | **28.8%** | **28.7%** | 0.0% |
| Leaps resolved | 77.2% | **91.1%** | **90.0%** | 52.2% |
| Unique max note | 64.7% | 56.4% | 59.4% | 37.1% |
| Unique min note | 49.4% | 51.9% | **58.3%** | **56.5%** |
| Notes in motif | 5.9% | **75.7%** | **73.8%** | **69.3%** |
| Notes in repeat motif | 0.007% | **0.11%** | **0.09%** | 0.01% |

Table 1: Statistics of music theory rule adherence based on 100,000 randomly initialized melodies generated by each model. The top half of the table contains metrics that should be near zero, while the bottom half contains metrics that should increase. Bolded entries represent significant improvements over the MLE baseline.

While the metrics indicate that the targeted behaviors of the RNN have improved, it is not clear whether the models have retained information about the training data. Figure 2a plots the average $\log p(a|s)$ as produced by the Reward RNN for melodies generated by the models every 100,000 training epochs; Figure 2b plots the average $r_T$. Included in the plot is an *RL only* model trained using only the music theory rewards, with no information about $\log p(a|s)$. Since each model is initialized with the weights of the trained MLE RNN, we see that as the models quickly learn to adhere to the music theory constraints, $\log p(a|s)$ falls from its initial point. For the RL only model, $\log p(a|s)$ reaches an average of -3.65, which is equivalent to an average $p(a|s)$ of approximately 0.026, or essentially a random policy over the 38 actions with respect to the distribution defined by the Reward RNN. Figure 2a shows that each of our models ($Q$, $\Psi$, and $G$) attain higher $\log p(a|s)$ values than this baseline, indicating they have maintained information about the data distribution, even over 3,000,000 training steps. The $G$-learning implementation scores highest on this metric, at the cost of slightly lower average $r_T$. This compromise between data probability and adherence to music theory could explain the difference in the $G$ model's performance on the music theory metrics in Table 1. Finally, we have verified that by increasing the $c$ parameter it is possible to train all the models to have even higher average $\log p(a|s)$, but found that $c = 0.5$ produced melodies that sounded better subjectively.

The question remains whether the RL-tutored models actually produce more pleasing melodies. The sample melodies used for the study are available here: goo.gl/XIYt9m;

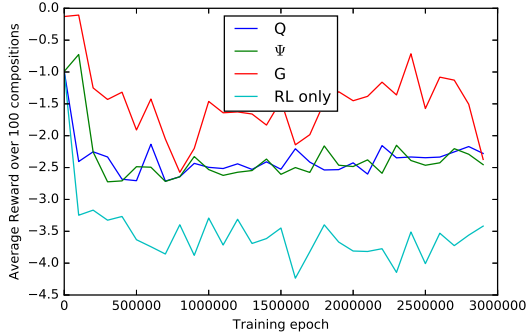

(a) Reward RNN reward: $\log p(a|s)$

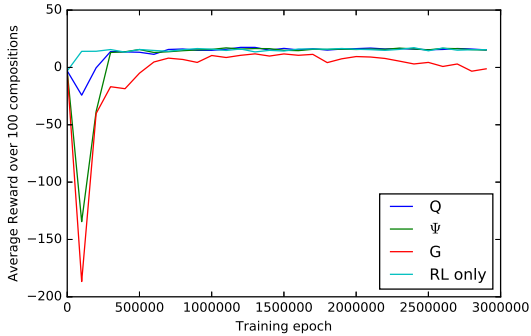

(b) Music theory reward

Figure 2: Average reward obtained by sampling 100 melodies every 100,000 training epochs. The three models are compared to a model trained using only the music theory rewards $r_T$.

we encourage readers to judge their quality for themselves. To more formally answer this question, we conducted a user study via Amazon Mechanical Turk in which participants were asked to rate which of two randomly selected melodies they preferred on a Likert scale. A total of 192 ratings were collected; each model was involved in 92 of these comparisons. Figure 3 plots the number of comparisons in which a melody from each model was selected as the most musically pleasing. A Kruskal-Wallis H test of the ratings showed that there was a statistically significant difference between the models, $\chi^2(3) = 109.480, p < 0.001$. Mann-Whitney U post-hoc tests revealed that the melodies from all three Sequence Tuner models ($Q$, $\Psi$, and $G$) had significantly higher ratings than the melodies of the MLE RNN, $p < .001$. The $Q$ and $\Psi$ melodies were also rated as significantly more pleasing than those of the $G$ model, but did not differ significantly from each other.

### 5.2. Discussion

Listening to the samples produced by the MLE RNN reveals that they are sometimes dischordant and usually dull; the model tends to place rests frequently, repeat the same note, and produce melodies with little variation. In contrast, the melodies produced by the Sequence Tutor models are more varied and interesting. The $G$ model tends to produce energetic and chaotic melodies, which include sequences of repeated notes. This repetition is likely because the $G$ policy as defined in Eq. 15 directly mixes $p(a|s)$ with the output of the $G$ network, and the MLE RNN strongly favours repeating notes. The most pleasant melodies are generated by the $Q$ and $\Psi$ models. These melodies stay firmly in key and frequently choose more harmonious interval steps, leading to melodic and pleasant melodies. However, it is clear they have retained information about the training data; for example, the sample q2.wav in the sample directory ends with a seemingly familiar riff.

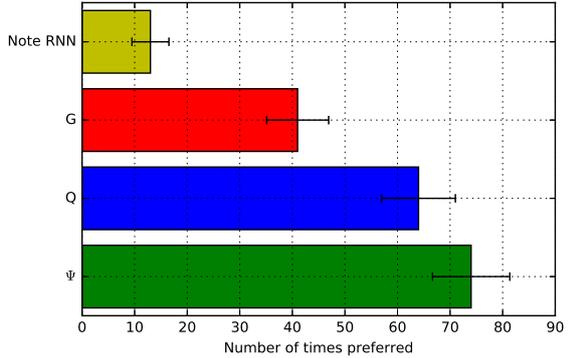

Figure 3: The number of times a melody from each model was selected as most musically pleasing. Error bars reflect the std. dev. of a binomial distribution fit to the binary win/loss data from each model.

While we acknowledge that the monophonic melodies generated by these models — which are based on highly simplistic rules of melodic composition — do not approach the level of artistic merit of human composers, we believe this study provides a proof-of-concept that encoding even incomplete and partially specified domain knowledge using our method can help the outputs of an LSTM adhere to a more consistent structure. The musical complexity of the songs is limited not just by the heuristic rules, but also by the simple monophonic encoding, which cannot represent the dynamics and expressivity of a musical performance. Although these melodies cannot surpass those of human musicians, attempting to train a model to generate aesthetically pleasing outputs in the absence of a better metric of human taste than log-likelihood is a problem of broader interest to the artificial intelligence community.

## 6. Experiment II: Computational Molecular Generation

As a follow-on experiment, we tested the effectiveness of Sequence Tutor for generating a higher yield of synthet-ically accessible drug-like molecules. Organic molecules can be encoded using the commonly used SMILES representation (Weininger, 1970). For example, amphetamine can be encoded as 'CC(N)Cc1ccccc1', while creatine is 'CN(CC(=O)O)C(=N)N'. Using this character encoding, it is straightforward to train an MLE RNN to generate sequences of SMILES characters; we trained such a model using the same settings as described above for the melody MLE RNN. However, only about a third of the molecules generated using this simple approach are actually valid SMILES encodings. Further, this approach does not directly optimize for metrics of molecule or drug quality. These metrics include: a) the water-octanol partition coefficient (logP), which is important in assessing the drug-likeness of a molecule; b) synthetic accessibility (SA) (Ertl & Schuffenhauer, 2009), a score from 1-10 that is lower if the molecule is easier to synthesize; and c) Quantitative Estimation of Drug-likeness (QED) (Bickerton et al., 2012), a more subjective measure of drug-likeness based on abstract ideas of medicinal aesthetics.

To optimize for these metrics, while simultaneously improving the percent yield of valid molecules from the RNN, we constructed a reward function that incentivizes validity, logP, SA, and QED using an open-source library called RDkit (http://www.rdkit.org/). Included in the reward function was a penalty for molecules with unrealistically large carbon rings (size larger than 6), as per previous work (Gómez-Bombarelli et al., 2016). Finally, after observing that the model could exploit the reward function by generating the simple molecule 'N' repeatedly, or 'CCCCC...' (which produces an unrealistically high logP value), we added penalties for sequences shorter than, or with more consecutive carbon atoms than, any sequence in the training data. Sequence Tutor was then trained using these rewards, the pre-trained MLE RNN, and similar settings to the first experiment, except with $\epsilon$-greedy exploration with $\epsilon = .01$, a batch size of 512, and discount factor $\gamma = .95$. For this experiment, we also made use of prioritized experience replay (Schaul et al., 2015) to allow the model to more frequently learn from relatively rare valid samples. A value of $c = 2.85$ led to a higher yield of valid molecules with high metrics, but still encouraged the diversity of generated samples.

### 6.1. Results and discussion

As the $\Psi$ algorithm produced the best results for the music generation task, we focused on using this technique for generating molecules. Table 2 shows the performance of this model against the original MLE model according to metrics of validity, drug-likeness, and synthetic accessibility. Once again, Sequence Tutor is able to significantly improve almost all of the targeted metrics. However, it should be noted that the Sequence Tutor model tends to produce simplistic molecules involving more carbon atoms than the MLE baseline; e.g. Sequence Tutor may produce 'SNCc1ccccc1', while the MLE produces 'C(=O)c1ccc(S(=O)(=O)N(C)C)c(Cl)c1', which is the reason for the Sequence Tutor model's lower QED scores. This effect is due to the fact that simple sequences are more likely to be valid, have high logP and SA scores, and carbon is highly likely under the distribution learned by the MLE model. A higher reward for QED and further improvement of the task-specific rewards based on domain knowledge could help to alleviate these problems. Overall, the fact that Sequence Tutor can improve the percentage of valid molecules produced as well as the logP and synthetic accessibility scores serves as a proof-of-concept that Sequence Tutor may be valuable in a number of domains for imparting domain knowledge onto a sequence predictor.

| Metric | MLE | Q |
| --- | --- | --- |
| Percent valid | 30.3% | **35.8%** |
| Mean logP | 2.07 | **4.21** |
| Mean QED | .678 | .417 |
| Mean SA penalty | -2.77 | **-1.79** |
| Mean ring penalty | -.096 | **-.001** |

Table 2: Statistics of molecule validity and quality based on 100,000 randomly initialized samples. Bolded entries represent significant improvements over the MLE baseline.

## 7. Conclusion and Future Work

We have derived a novel sequence learning framework which uses RL to correct properties of sequences generated by an RNN, while maintaining information learned from MLE training on data, and ensuring the diversity of generated samples. By demonstrating a connection between our sequence generation approach and KL-control, we have derived three novel RL-based methods for optimizing sequence generation models. These methods were empirically compared in the context of a music generation task, and further demonstrated on a computational molecular generation task. Sequence Tutor showed promising results in terms of both adherence to task-specific rules, and subjective quality of the generated sequences.

We believe the Sequence Tutor approach of using RL to refine RNN models could be promising for a number of applications, including the reduction of bias in deep learning models. While manually writing a domain-specific reward function may seem unappealing, that approach is limited by the quality of the data that can be collected, and besides, even state-of-the-art sequence models often fail to learn all the aspects of high-level structure (van den Oord et al., 2016; Graves, 2013). Further, the data may contain hidden biases, as has been demonstrated for popular language models (Caliskan-Islam et al., 2016). In contrast to relying solely on possibly biased data, our approach allows for encoding high-level domain knowledge into the RNN, providing a general, alternative tool for training sequence models.

ACKNOWLEDGMENTS

This work was supported by Google Brain, the MIT Media Lab Consortium, and Canada's Natural Sciences and Engineering Research Council (NSERC). We thank Greg Wayne, Sergey Levine, and Timothy Lillicrap for helpful discussions on RL and stochastic optimal control and Kyle Kastner and Tim Cooijmans for valuable insight into training RNNs.